%% file: TRA-v2.0.tex
\documentclass[journal]{IEEEtran}

\input{usercommands}
\title{Design and control of a soft, shape-changing, crawling robot}

\begin{document}
%\title{Locomotion of four-limbed soft-bodied robot}
\author{Vishesh~Vikas,~Paul~Templeton, and~Barry~Trimmer% <-this % stops a space
\thanks{ Vishesh Vikas, Barry Trimmer and Paul Templeton are with Neuromechanics and Biomimmetic Devices Lab at Tufts University, Medford MA 02155} 
}
\maketitle
%%%%%%%%%%%%%%%%%%%%%%%%%%%%%%%%%%%%%%%%%%
\begin{abstract}
Soft materials have many important roles in animal locomotion and object manipulation. In robotic applications soft materials can store and release energy, absorb impacts, increase compliance and increase the range of possible shape profiles using minimal actuators. The shape changing ability is also a potential tool to manipulate friction forces caused by contact with the environment. These advantages are accompanied by challenges of soft material actuation and the need to exploit frictional interactions to generate locomotion. Accordingly, the design of soft robots involves exploitation of continuum properties of soft materials for manipulating frictional interactions that result in robot locomotion. The research presents design and control of a soft body robot that uses its shape change capability for locomotion. The bioinspired (caterpillar) modular robot design is a soft monolithic body which interacts with the environment at discrete contact points (caterpillar prolegs). The deformable body is actuated by muscle-like shape memory alloy coils and the discrete contact points manipulate friction in a binary manner. This novel virtual grip mechanism combines two materials with different coefficients of frictions (sticky-slippery) to control the robot-environment friction interactions. The research also introduces a novel control concept that discretizes the robot-environment-friction interaction into binary states. This facilitates formulation of a control framework that is independent of the specific actuator or soft material properties and can be applied to multi-limbed soft robots. The transitions between individual robot states are assigned a reward that allow optimized state transition control sequences to be calculated. This conceptual framework is extremely versatile and we show how it can be applied to situations in which the robot loses limb function.
\end{abstract}

\section{Introduction}
Soft materials in nature allow animals to interact and adapt to uncertain and dynamically changing environments. The extraordinary versatility of soft structures is visible in the locomotion of worms and caterpillars or in manipulation by octopus and elephant trunks. Soft materials also play a major role in the performance of animals with stiff articulated skeletons by providing joint compliance, elastic energy storage, impact resistance and reliable limb-to-surface contact regimens \cite{vogel_comparative_2013}. Motivated by 	the robustness and adaptability of living systems there is increased interest in developing bio-inspired soft robots \cite{wang_locomotion_2014,lin_goqbot:_2011,seok_peristaltic_2010,%
laschi_design_2009,hannan_kinematics_2003,quinn_parallel_2003}. Introducing softness into robot designs makes them safe for operation, increases their range of movement and improves their performance in complex environments. However, soft materials  also create design challenges that are different from those of traditional ``rigid/hard'' engineering. Consequently, it is essential to explore and elaborate a novel set of design and control principles that help to bridge a gap between hard and soft engineering approaches \cite{pfeifer_challenges_2012}. %

Terrestrial locomotion of robots using undulatory motion \cite{hirose_biologically_2004,chirikjian_kinematics_1995,%
ostrowski_geometric_1996,wright_design_2007,maladen_undulatory_2009}, compliant limbs \cite{quinn_parallel_2003,saranli_rhex:_2001} and soft bodies \cite{shepherd_multigait_2011,wang_locomotion_2014,%
umedachi_highly_2013} have been important topics in the field of bioinspired robotics.
%%
%Terrestrial locomotion of robots utilizing undulatory motion \cite{hirose_biologically_2004,chirikjian_kinematics_1995,%
%ostrowski_geometric_1996,wright_design_2007,maladen_undulatory_2009} and compliant limbs \cite{quinn_parallel_2003,saranli_rhex:_2001} have been researched. Bioinspired bodied robot capable of terrestrial locomotion have been also been researched \cite{shepherd_multigait_2011,wang_locomotion_2014,%
%umedachi_highly_2013}. 
This research presents the design of a robot inspired by caterpillars using a soft deformable body with discrete contact points between the robot and the environment. The platform is modular, allowing it to be expanded to test problems of increasing complexity. In this paper we present some design considerations for a bioinspired robot, introduce a novel state transition control for the soft robot and conclude with a discussion of the experimental results and significance of our findings. 
\begin{figure}
\centering
\includegraphics[scale=0.5]{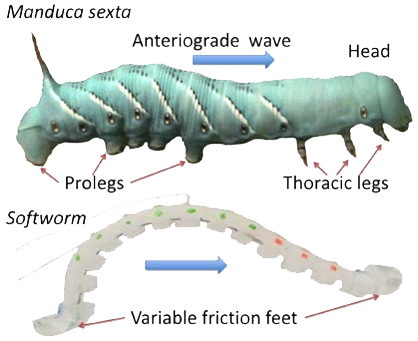}
\caption{Caterpillar-inspired design of a modular soft robot with a deformable body. The soft body is actuated by muscle-like shape memory alloys coils and the friction is manipulated at discrete points using virtual grip mechanisms.}
\label{Fig:CaterpillarRobot}
\end{figure}
%%%%%%%%%%%%%%%%%%%%% ROBOT BODY %%%%%%%%%%%%%%%%%%%%%%%%%%%%%%%%%%%%
\section{Design of modular soft}
\subsection{Design challenges}
The two key challenges to designing soft robots capable of locomotion are \textit{frictional interactions} and \textit{soft actuation}. 

\textit{Frictional interactions.} A major factor in effective terrestrial locomotion is the changing contact between the body and its environment. The forces required to initiate or maintain differential movement between interacting surfaces are often dominated by friction. Animals have evolved a variety of mechanisms to exploit these forces including directionally sensitive friction \cite{autumn_frictional_2006}, adhesives \cite{creton_sticky_2007}, structures that exploit asperities in different size ranges \cite{gorb_attachment_2001} and deployable or retractable grippers \cite{mezoff_biomechanical_2004,belanger_combined_2000}. In general, locomotion results from optimization of frictional forces by minimizing friction at one end while maximizing it at the other \cite{radhakrishnan_locomotion:_1998}. The gripper-like variable friction mechanism, discussed in Section \ref{Subsec:RobotEnvInteraction}, achieves a similar effect using two materials with different friction coefficients that contact the ground as a function of the changing robot shape. 

\textit{Soft actuation.} Actuation of soft robots remains a challenge because most electromagnetic systems are made of hard materials \cite{kim_soft_2013}. Alternative systems such as dielectric elastomeric actuators (DEAs) have been explored \cite{ohalloran_review_2008,carpi_biomedical_2009}, but these require high voltages for actuation and, without a rigid frame, produce very low stress \cite{cianchetti_new_2009}. Other flexible actuators include pressurized liquid or air \cite{daerden_pneumatic_2002,wehner_experimental_2012,%
shepherd_multigait_2011} and cable-driven systems \cite{laschi_design_2009}. The platform described here uses shape memory alloy (SMAs) coils to amplify the overall strain \cite{lin_goqbot:_2011,laschi_design_2009,kim_soft_2013,%
umedachi_highly_2013}. These are activated through resistive heating which causes them to contract. Analogous to natural muscles, re-extension requires an external force which is provided by another SMA coil or intrinsic elasticity of the deformed body. Typically, the soft actuators are embedded inside the soft body to provide it the dual actuator-structure function. Apart from type of actuation, the design of actuation also needs to ensure controlled deformation of the soft material body. 

\subsection{Robot body and actuation}
With these design features in mind we introduce a new platform for soft robotics research. The soft robot design incorporates structural control, friction manipulation, soft actuation using minimum actuators and a monolithic body. The design is modular and can be used to build robots with increased structural complexity. This will facilitate systematic analysis of robust control strategies for highly deformable robots. %
The basic module of this robot is a monolithic structure with actuators that can be combined with others to create ``multi-limbed'' soft devices. The design choices for the robot are \textit{the structure and fabrication} - the soft material(s), the shape of the robot and fabrication technique, \textit{actuation }- the type, number and placement of the actuators, and the \textit{friction manipulation} mechanism. Because the robot is a continuum system, these design features are intricately coupled.
Soft linear animals like caterpillars locomote by changing their body shape and manipulating friction at finite points using their passive gripping prolegs \cite{van_griethuijsen_locomotion_2014} (Figure \ref{Fig:CaterpillarRobot}). These animals are the design inspiration for the soft robot body and gripper-like friction manipulation mechanism which utilizes differential friction to facilitate locomotion.
%%%%%%%%%%%%%%%%%%%%%%%%%%%%%%%%%%%%%%%%%%%%%%%%%% 

\begin{figure}
\centering
\includegraphics[width=\columnwidth]{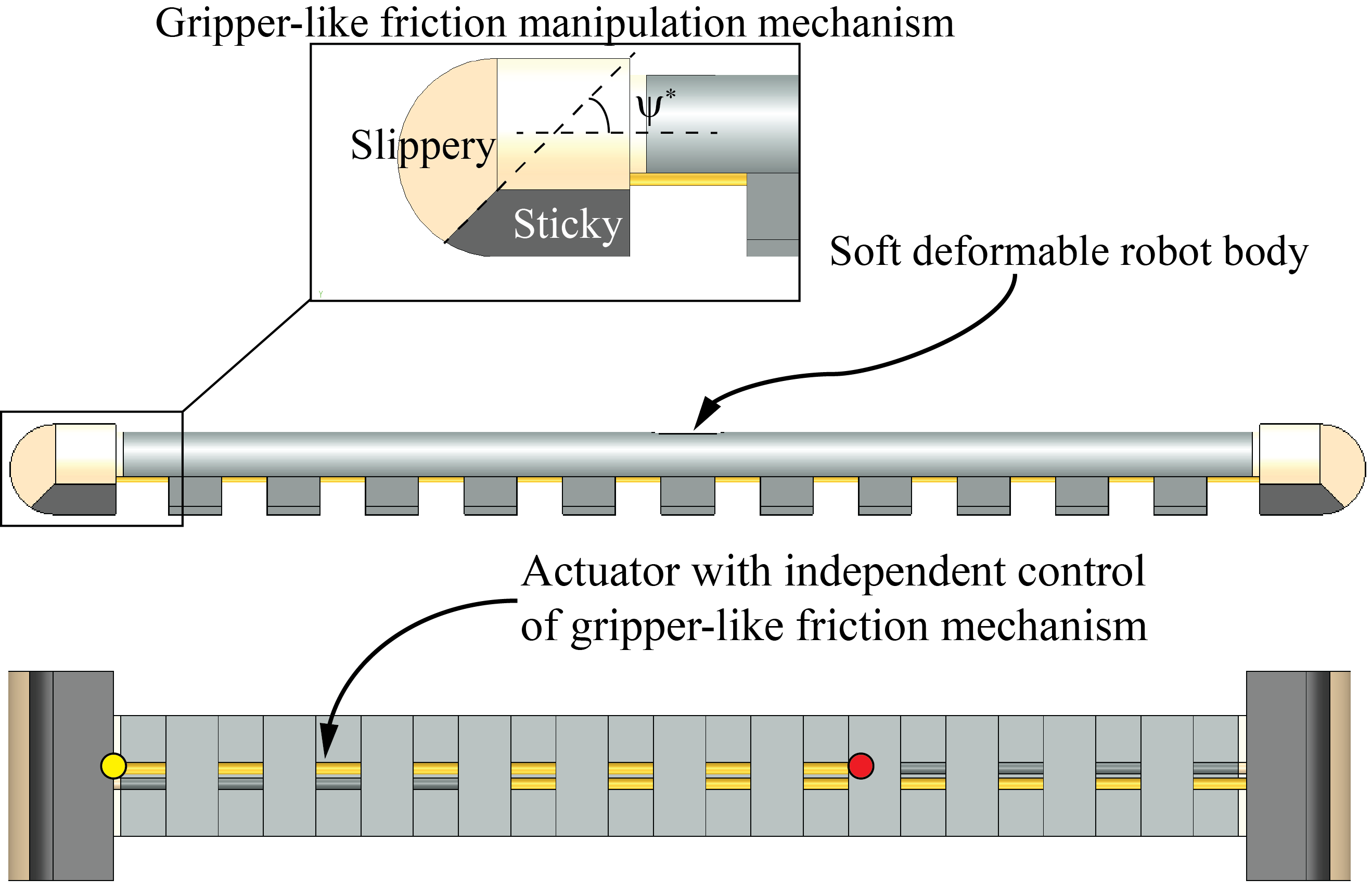}
\caption{The top and side views of the soft robot. The body of the robot is made out of soft material. The $80mm$ long soft robot body is controlled using two overlapping SMA coils. The virtual grip variable friction mechanisms allow controlled manipulation of friction.} 
\label{Fig:overallassembly}
\end{figure}

\textit{Structure design and fabrication.} The robot body is made from a deformable material with horizontal ribs (Figure \ref{Fig:overallassembly}). The ribs facilitate bending in the desired direction and cooling of the temperature-dependent actuators. The whole device is printed on a multi-material printer  (Connex 500\texttrademark) \cite{_http://www.stratasys.com/_????} using TangoPlus\texttrademark\ (Shore A Hardness)  as the soft, rubber-like material and VeroClear\texttrademark\ (Shore D Harness) for the hard material. The robot body is 80 mm long and weighs between 3 to 3.6 gm (Figure \ref{Fig:overallassembly}).

\begin{figure}
\centering
\subfloat[Actuation pattern of a single SMA. $T_{active}$ indicates the time period the SMA is actuated and $(T_{cycle}-T_{active})$ is the time that allows for the SMA to cool. Strength ($S$) is proportional to the current supplied to the actuator.]{
\includegraphics[width=0.75\columnwidth]{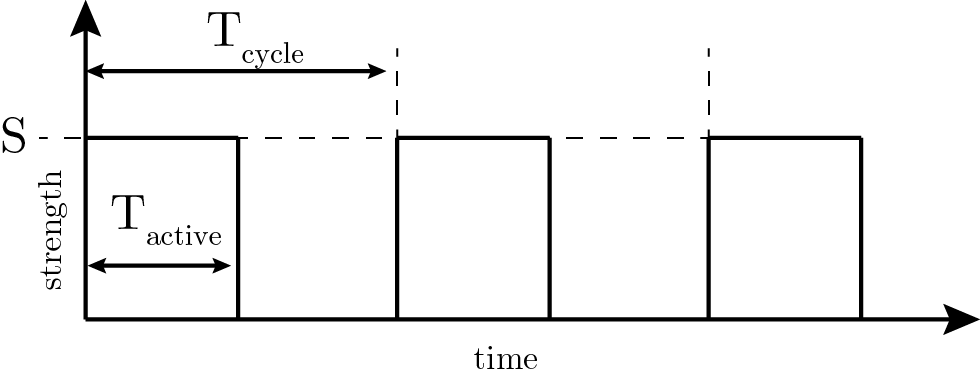}%
\label{Fig:SingleSMA}
}\\
\subfloat[Independent actuation of two SMA actuators increases the number of control parameters to seven - three for each SMA actuator (strength, active time, cycle time) and one for time gap between actuation of each SMA, referred to as phase $\phi_{12}$.]{
\includegraphics[width=0.75\columnwidth]{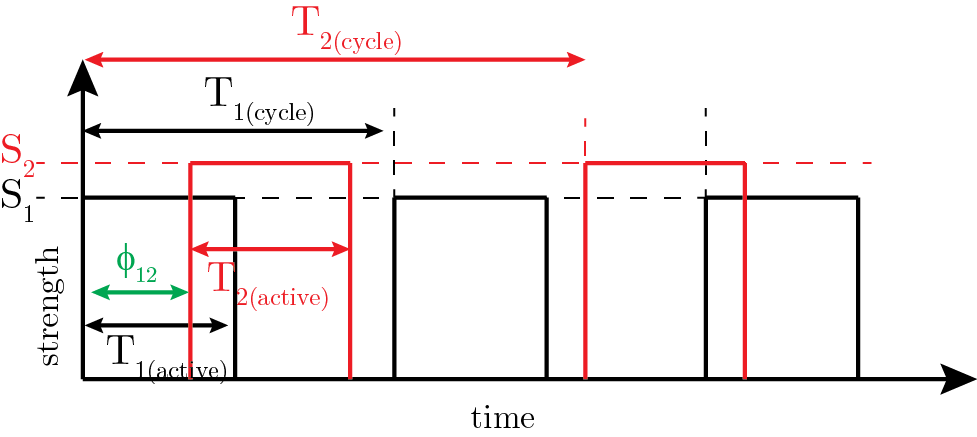}
\label{Fig:TwoSMA}
}
\caption{Shape memory alloy based actuation pattern for one and two actuators.}
\label{Fig:actuation_SMA}
\end{figure}

\textit{Actuation.} The robot is actuated using two shape memory alloy (SMA) coils \cite{_http://www.toki.co.jp/biometal/english/contents.php_????} that are threaded through open channels inside the robot above the mid-line of the robot body resulting in concave shape bending when they are actuated (Figure \ref{Fig:overallassembly}). The location of SMA attachment points and the resulting overlap may be varied for different robots with the condition that each actuator can independently control the friction mechanism at each end of the robot. Each actuator is pre-stretched when it is installed and is activated by using current pulses that heat the SMA causing the coil to shorten. Elastic forces in the body materials restore the original actuator length when it is allowed to cool. Activation of a single SMA actuator is periodic where the strength of the actuation current pulse ($S$, proportional to actuation force), time of actuation ($T_{active}$) and time of periodic cycle ($T_{cycle}$) can be varied (Figure \ref{Fig:SingleSMA}). The SMA actuators are inconsistent over time, mainly because of the non-uniform transitions (due to cooling rate, etc.) of crystals between the austenite and martensite phases \cite{wayman_shape_1993}.
%%%%%%%%%% 
%%%%%%%%%%%%%%%%%%%%%%%%%%%%%%%%%%%%%%%%%%%%%%%%%%%%%%%%%%%%%%%%
The independent control of two SMA actuators is done using seven (7) parameters - the strength, time of actuation, time cycle for each SMA and the time gap ($\phi_{12}$) between the respective periodic actuation cycles as illustrated in Figure \ref{Fig:TwoSMA}.
%When two SMA actuators are involved, to facilitate independent control, the time gap between the periodic actuation of first and second SMA needs to be controlled and is referred to as phase - $\phi_{12}$ as indicated in Figure \ref{Fig:actuation_SMA}b.
%%%%%%%%%%%%%%%%%%%%%%%%%%%%%%%%%%%%%%%%%%%%%%%%%%%%%%%%%%%%%%%%
%%%%%%%%%%

\begin{figure}
\centering
\includegraphics[width=0.75\columnwidth]{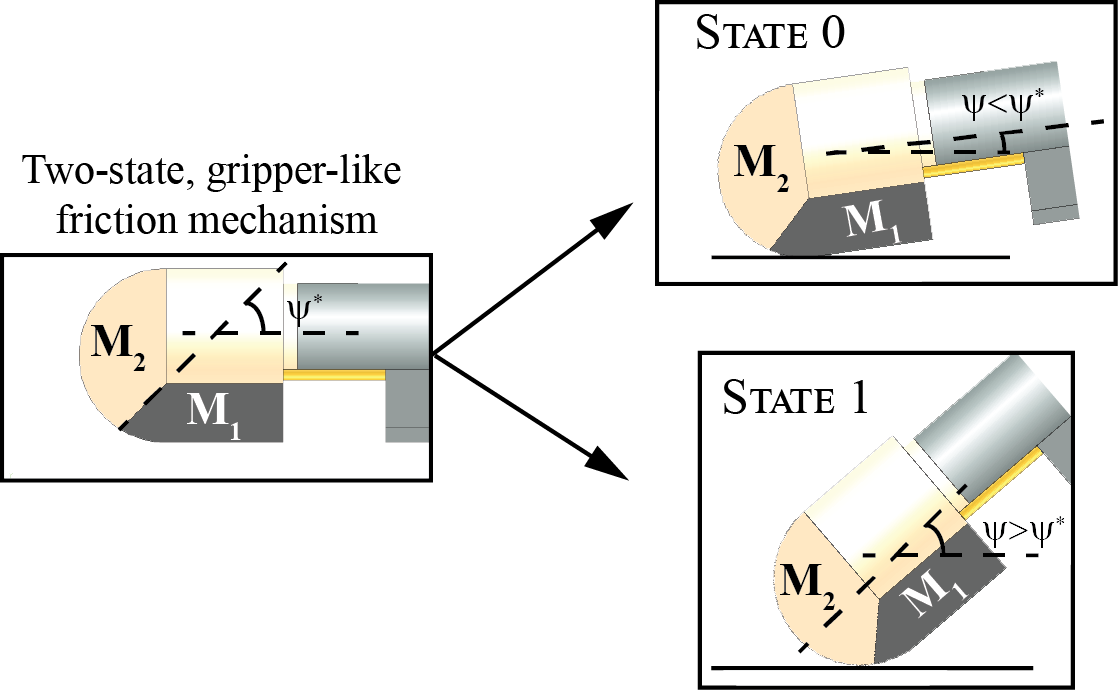}
\caption{The variable friction mechanism consists of two materials ($\mathrm{M}_1$ and $\mathrm{M}_2$) which have different friction coefficients. The angle between the body tangent at the end and the surface horizontal is referred to as the contact angle $\psi$. The coefficient of friction changes after the critical contact angle ($\psi^*$). The binary states $0,\ 1$ directly relate to the material in contact i.e. friction.} 
\label{Fig:frictionleg}
\end{figure}

\begin{figure}
\centering
\subfloat[]{\includegraphics[height=80pt]{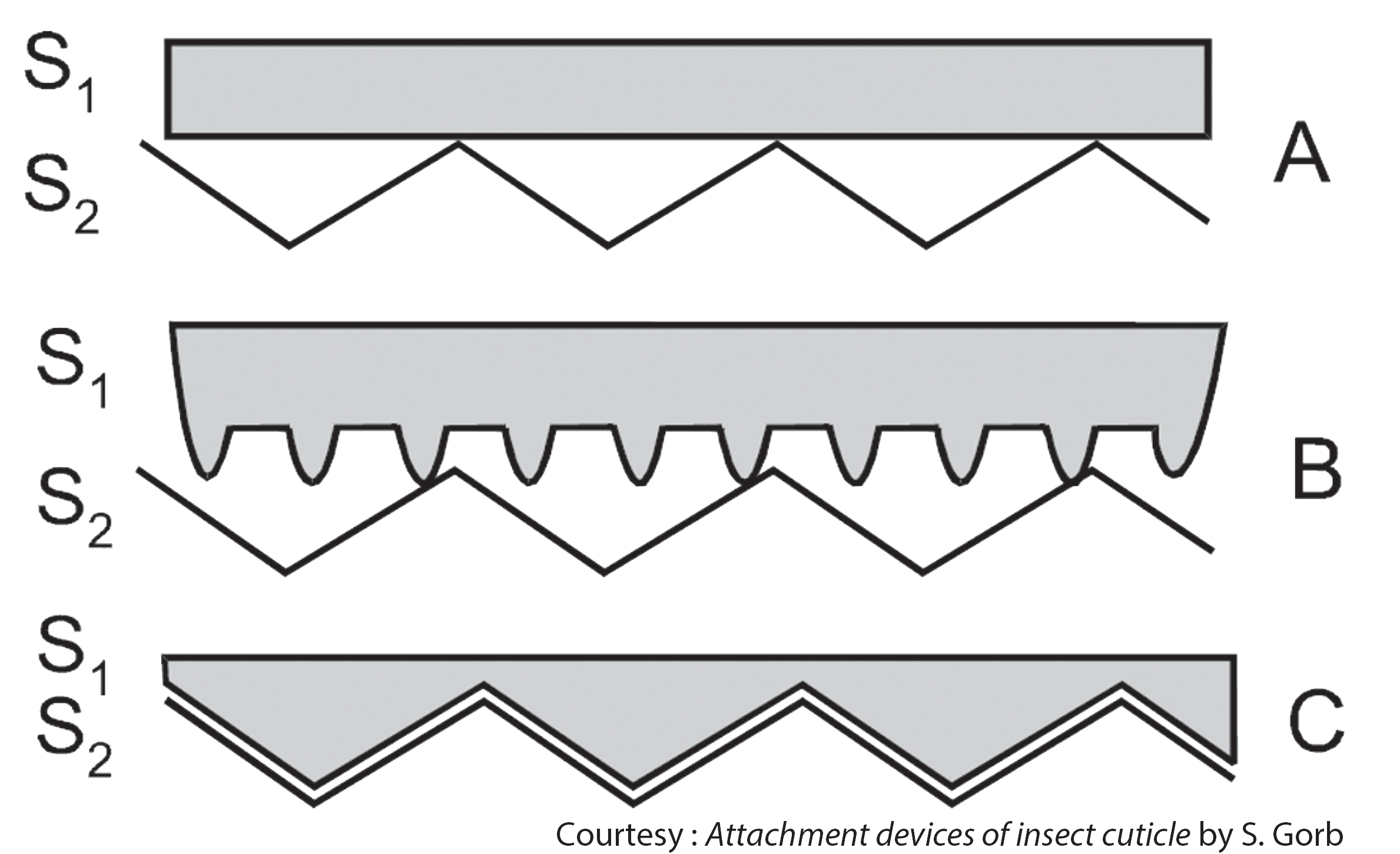} }
\subfloat[]{\includegraphics[height=80pt]{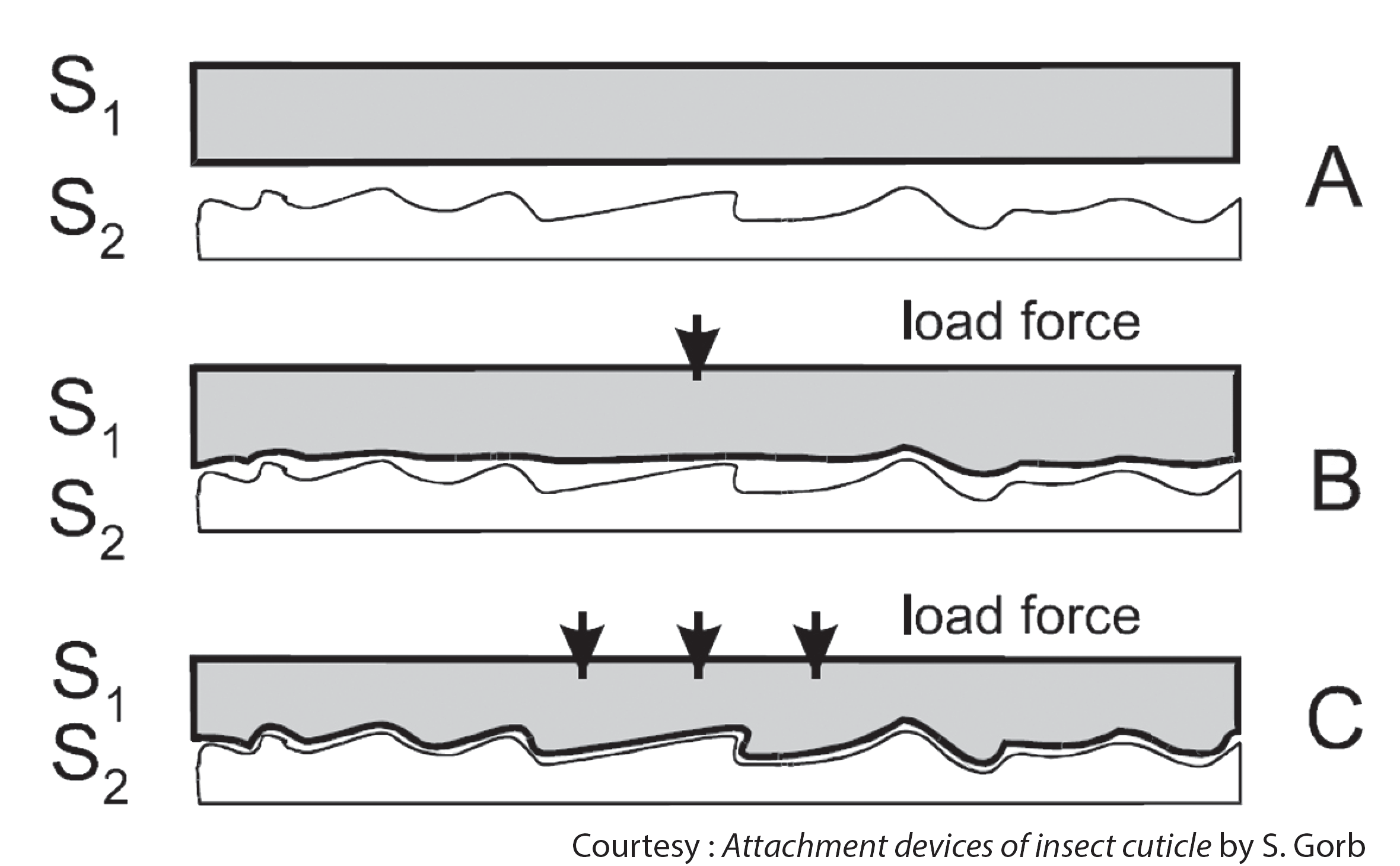} }
\caption{(a) Friction depends on profile of both hard surfaces. $\mathrm{S}_1$ and $\mathrm{S}_2$ have irregularities of different dimensions in (A) and (B), thus, there will be considerably lower friction as to when the irregularities are of similar dimension (C) (b) When soft material ($\mathrm{S}_1$) and hard material ($\mathrm{S_2}$) are in contact, the flow of the soft material conform to asperities of the hard material resulting in increase of friction force. \cite{gorb_attachment_2001}}
\label{Fig:softandhard}
\end{figure}

\textit{Friction manipulation mechanism.} \label{Subsec:VFLM}
%%%%%%%%%%%%%%%%%%%%%%%%%%%%%%%%%%%%%%%%%%%%%%%%%%%%%%%%%%%%%%%
%%%%%%%%%%%%%%%%%%%%%%%%%%%%%%%%%%%%%%%%%%%%%%%%%%%%%%%%%%%%%%%
Frictional force arises from the interaction of irregularities between surfaces in contact \cite{gorb_attachment_2001,asbeck_scaling_2006} and depends on the material and its texture. The change of robot shape (bending) upon actuation changes the angle between the body tangent at the end of the beam-like robot body and the surface horizontal which is referred to as the contact angle $\psi$. This is the basis of the shape dependent friction mechanism (Figure \ref{Fig:frictionleg}). The contact surface is made from two different materials $M_1,\ M_2$ with different coefficients of frictions. As the contact angle $\psi$ changes about the critical contact angle $\psi^*$ the friction changes from one value to the other. This two-material differential friction mechanism is similar to a biological gripper such as the caterpillar proleg that is on (very high friction) or off (zero friction). Hence, the friction mechanism is identified by binary states ($S$) such that
\begin{equation} \label{Eqn:StateS}
S = \left\{ \begin{array}{c @{\ for\ } c}
0 & (\psi-\psi^*)<0\\
1 & (\psi-\psi^*)\geq 0
\end{array}
\right.
\end{equation}
This friction mechanism makes use of the different interactions exhibited by soft and hard materials on surfaces of varying roughness (Figure \ref{Fig:softandhard}). When two hard materials are pressed together, friction is largely determined by the effective area of contact which itself depends on how well the surface irregularities interlock. This is a function of the size and match of asperities in the two materials (Figure \ref{Fig:softandhard}a). However, during interactions between hard and soft surfaces the soft/flexible material has the ability to flow and conform to asperities of the harder material surface, thus, increasing friction (Figure \ref{Fig:softandhard}b). In this case the friction depends far more on the applied load than it would for two hard surfaces \cite{gorb_attachment_2001}.
%%%%%%%%%%%%%%%%%%%%%%%%%%%%%%%%%%%%%%%%%%%%%%%%%%%%%%%%%%%%%%%%%%%%%%%%%%%%%%%%%%%%%%%%%%%%%%%%%%%%%%%%%%%%%%%%%%%%%%%%%%%%%%
%%%%%%%%%%%%%%%%%%%%%%%%%%%%%%%%%%%%%%%%%%%%%%%%%%%%%%%%%%%%%%%
Hence, in the present case for the same load applied, the soft material is stickier with higher coefficient of friction compared to the hard material, thus, providing differential friction. The mechanism works as follows - as the SMA coil is electrically actuated, the consequent bending of the robot results in change of the contact angle. When the contact angle $\psi$ exceeds the critical contact angle $\psi^*$, the contact material changes from $M_1$ to $M_2$. Upon deactivation (no current flow), the SMA coils cool and the intrinsic elasticity of the bent body results in straightening of the robot, equivalently, decrease in contact angle. Now, when the contact angle is less than the critical contact angle, the contact material changes to $M_2$. So, this gripper-like friction manipulation mechanism allows for the friction force acting on the robot to change with its shape. This design can be interpolated to multi-state, multi-material (with different friction coefficient) friction manipulation mechanisms. 

\subsection{Robot-environment interaction} \label{Subsec:RobotEnvInteraction}
\begin{figure}
\centering
%\ifconvert
\includegraphics[width=0.75\columnwidth]{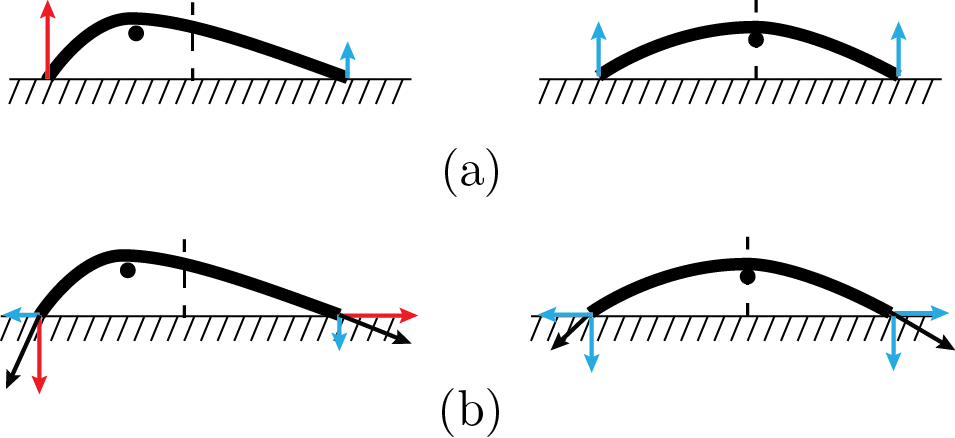}
%\else
%\includegraphics[width=\columnwidth]{figures/shape_friction.eps}
%\fi
\caption{Symmetric and non-symmetric bending of the monolithic soft body robot. (a) SMA contraction - due to shifted center of mass (circle) in non-symmetric bending, the normal force at one end is more than the other. (b) SMA relaxation - the push force parallel to the surface is more at one as compared to the other for non-symmetric robot bending.}
\label{Fig:shapefriction}
\end{figure}

The robot interacts with the environment via the discrete contact friction manipulation mechanisms. This interaction is complicated to analyze and model e.g. quasi-static analysis of this frictional interaction makes it dependent upon the coefficient of friction (material $M_1,\ M_2$) and the normal force at each friction mechanism. The normal force is coupled to changes in shape of the robot which depends on the placement, activation of the actuators. As a result, non-symmetric and symmetric bending of the robot creates different normal and frictional forces for contraction (Figure \ref{Fig:shapefriction}a) and relaxation (Figure \ref{Fig:shapefriction}b).
However, the robot-environment interaction at finite contact points and discretization of the contact behavior (Equation \ref{Eqn:StateS}) allows the robot behavior to be defined. 

The \textit{robot behavior} is defined as the states of the two friction mechanisms and is written as $(S_1\ S_2)$ where $S_i$ corresponds to state of the $i$th friction mechanism. Exhaustively, the four robot states in the present case are $\{(00), (01), (10), (11)\}$. Discretization of the robot-environment interaction using these robot states is instrumental for the control system design. 
%%%%%%%%%%%%%%%%%%%%
%%%%%%%%%%%%%%%%%%%%%%%%%%%%%%%%%%%%%%%%%%%%%%%%%%%%%%%%%%%%%%%%%%%%
%%%%%%%%%%%%%%%%%%%% CONTROL %%%%%%%%%%%%%%%%%%%%%%%%%%%%%%%%%%%%%%%
\section{Soft robot control}
Control of soft material robots is typically performed by continuum modeling techniques \cite{ostrowski_geometric_1996,chirikjian_kinematics_1995,%
webster_design_2010,walker_continuum_2011,cianchetti_design_2011} and finite element methods \cite{duriez_control_2013}. A particular challenge for these approaches in terrestrial locomotion is that interaction of the robot with the environment is difficult to model and simulate.  However, the discretization of finite point robot-environment interaction discussed in Section \ref{Subsec:VFLM}, \ref{Subsec:RobotEnvInteraction} facilitates development of a robot state-based control framework. This framework indirectly models the robot-environment interaction and helps in calculation of optimal control sequences for locomotion.
\subsection{State transition and control framework}
As discussed in Section \ref{Subsec:RobotEnvInteraction}, the robot can exist in one of four states. The transition from one state to another results in linear displacement of the center of mass (termed as \textit{reward}) - forward, backward or none that is discretized as $+1$, $-1$ or $0$. The result of a single state to state transition is stored inside a \textit{state transition reward matrix} $T \in \mathcal{R}^{2 \times 2}$. The element $T_{ij}$ represents the displacement for a transition from state $\mathrm{dec2bin}(i-1)$ to $\mathrm{dec2bin}(j-1)$ e.g. $T_{13}$ represents the discretized displacement when the robot transitions from state $(00)$ to $(10)$. Rather than direct modeling of surface interactions, the transition matrix represents relative differences in friction, a critical aspect of locomotion. %
This framework is referred to as the model-free control framework. It is expected to have general applications with the following advantages 
\begin{enumerate}
\item \textit{Material and actuator independence.} The transition from one state to another is only dependent on the critical contact angle ($\psi^*$). The robot state can be measured using an angular feedback sensor (e.g. MEMS inertial sensor) without the need to model the actuator (SMA, motor-tendon, pneumatic, etc) or even the specifics of the body material (rubber, PDMS, TangoPlus\texttrademark)
\item \textit{Friction manipulation mechanism independence.} The framework is applicable to scenarios where the friction manipulation can be discretized into finite number of behaviors such as a gripper (zero, infinite friction) or a unidirectional roller. Furthermore, the number of discretized behaviors is not restricted to two but can be applied to friction mechanisms made of multiple materials (two or more critical contact angles).
\item \textit{Simplification of control parameters.} Using this framework, the control parameters for $N$ friction mechanisms decrease from $4N-1$ to $N$ (number of critical contact angles). For the robot presented here, the number of control parameters decreases from $7$ (Figure \ref{Fig:actuation_SMA}) to $2$. Furthermore, changes in the SMA actuation properties can be easily compensated using the discretization process.
\item \textit{Adaptability to changing environment.} The state transition rewards result from the robot-environment interaction which provides a relatively straightforward mechanism to adapt to changing environments. The adaptability is expected to be achieved through a layer of intelligence that learns the state transition rewards as environment changes.
\item \textit{Calculation of optimized control sequences.} State transition rewards represent the transition between two individual states. However, it is possible to calculate optimal control sequences by optimizing cost functions. This is discussed in Section \ref{Subsection:ControlSequence}.
\item \textit{Extendible to multiple limbs/actuators.} The framework is extendible to multi-limbs/actuators. %As an example, a similar soft robot with $N+1$ limbs performing linear locomotion with binary state friction manipulation, the transition matrix will be of size $2^N\times 2^N$. The control framework will evolve for planar locomotion where the result of robot environment interaction can be viewed as a vector - change in position and orientation. The described soft robot can be visualized to comprise of two limb each actuated by an actuator. 
\end{enumerate}
%
%%%%%%%%
\subsection{Control sequences and speed} \label{Subsection:ControlSequence}
A control sequence is defined as a sequence of state transitions $S(t)$ for $t=1,2,\cdots,N$. The resulting translation reward $J_x$ for the given sequence is written as
\begin{equation} \label{Eqn:SeqReward}
J_x \left(\{S(t)\}\right)=\sum \limits_{t=1}^N T_{S(t-1),S(t)} 
\end{equation}
The calculation of optimal periodic control sequence $S_{x}^*$ for maximum translation in $+X$ direction can be calculated as
\begin{equation}
\label{Eqn:S_x}
S_{x}^* = \max \limits_{S(t)\ s.t.\ S(0)=S(N)}\left( J_x \right)
\end{equation}
with the constraint $N\leq l_{max}$ where $l_{max}$ is the maximum length of the sequence. For the research experiment $l_{max}=4$. Here it is important to note that the \textit{speed} of the robot locomotion directly depends upon its ability to transition from one state to another which dictates the speed of implementation of the control sequences. 
\section{Experiments}

The model-free control framework was used in experiments  to explore how soft robots with different friction arrangements affected the optimal locomotion sequences. The state transition matrix is unique to each robot as it corresponds to the unique robot-environment interaction and needs to be learned. This state transition matrix is critical for calculating the optimal control sequence (Equations \ref{Eqn:SeqReward}, \ref{Eqn:S_x}). For this initial study, the matrices are manually learned from visual feedback.
\newcommand{\design}[1]{$D_{#1}$}
\newcommand{\legwidth}{50pt}
\newcommand{\AddPhantomMinusIfNeeded}[1]{%
\IfDecimal{#1}{% Is a decimal, so if not negative add a \phantom{-}%
    \IfBeginWith{#1}{-}{\ensuremath{#1}}{\ensuremath{\phantom{-}#1}}%
    }{%
        \ensuremath{\phantom{-}}#1% Not a decimal number so just leave it alone
    }%
}%
\newcolumntype{L}{>{\collectcell\AddPhantomMinusIfNeeded}{l}<{\endcollectcell}}

Using soft and hard materials in the variable friction mechanism allows two order-dependent designs referred to as \design{1} ($M_1$ is soft, $M_2$ is hard) and \design{2} ($M_1$ is hard, $M_2$ is soft). Consequently, three possible combinations are explored - two symmetric cases with \design{1} at both ends (robot $R_1$) and \design{2}  at both ends (robot $R_2$). The third non-symmetric design has \design{1} at one end and \design{2} at the other end (robot $R_3$). The Table \ref{Tab:TransitionMats} records the robot type and transition matrix associated with the corresponding robot where soft, hard materials are indicated by dark grey and pale yellow colors respectively. % 
\begin{table}
\centering
\begin{tabular}{|c|C{\legwidth}|C{\legwidth}|c|}
\midrule  &
{\normalsize Left} & {\normalsize Right} & {\normalsize Transition Matrix}\\
\midrule \hline $R_1$ &
\includegraphics[width=\legwidth]{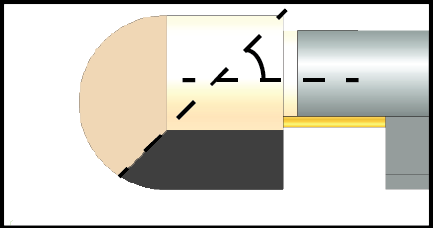} & %
\includegraphics[width=\legwidth]{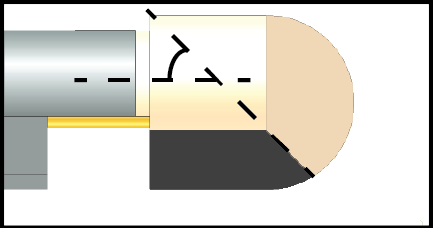} & %
$\left[\begin{array}{LLLL}
0 & -1 & 1 & 0\\
1 & 0 & 0 & -1\\
-1 &0 & 0 & 1\\
0 & 0 & 0 & 0
\end{array}\right]$\\ \midrule $R_2$ &
\includegraphics[width=\legwidth]{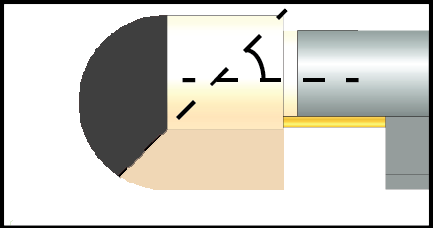} & %
\includegraphics[width=\legwidth]{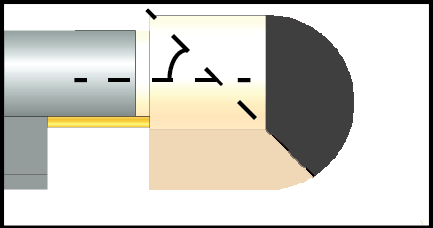} & %
$\left[ \begin{array}{LLLL}
0 & 1 & -1 & 0\\
-1 & 0 & 0 & 1\\
1 & 0 & 0 & -1\\
%0 & -1 & 1 & 0
0 & 0 & 0 & 0
\end{array}
\right]$\\ \midrule $R_3$ &
\includegraphics[width=\legwidth]{figures/Des1Left.png} & %
\includegraphics[width=\legwidth]{figures/Des2Right.png} & %
$\left[ \begin{array}{LLLL}
0 & -1 & 1 & 1\\
1 & 0 & 0 & -1\\
-1 &-1 & 0 & -1\\
-1 & 1 & 0 & 0
\end{array}
\right]$ \\
\hline
\end{tabular}
\caption{Transition matrices corresponding to the three combinations. The two hard, soft materials in the virtual grip friction mechanism are illustrated by dark gery and pale yellow colors respectively. The first two rows corresponds to identical friction mechanisms at both ends while the third row corresponds to non-symmetrical arrangement.}
\label{Tab:TransitionMats}
\end{table}
The calculation of optimal locomotion sequences in $+X, -X$ directions using Equation \ref{Eqn:S_x} for the three robots resulted in two periodic control sequences - $C_{T1} = \{(00) \rightarrow (10) \rightarrow (01) \rightarrow (00)\}$ and $C_{T2} = \{(00) \rightarrow (01) \rightarrow (10) \rightarrow (00)\}$. These correspond to propagation of anteriograde wave and retrograde waves respectively. 
The transition matrices for symmetrical designs are transpositions of one-another - hence the similarity in locomotion sequences. With the same execution time for both four state control sequences, movement is identical in both directions ($+4$ and $-4$) using $C_{T1},\ C_{T2}$ control sequences for $R_1$, $R_2$ robots (see supplemental video). %
The non-symmetrical robot $R_3$ displays different behavior - moving more slowly in the forward $+X$ direction using an anteriograde $C_{T1}$ sequence and faster in the backward direction using a retrograde $C_{T2}$ control sequence. Because the critical contact angle only is relevant for controlling locomotion, the state ($\psi>\psi^*$ or $\psi<\psi^*$) can be monitored visually very conveniently.

\textit{Loss of limb/actuation.} This control scheme should be robust to changes in the physical structure such as limb-loss. To test this scenario, the $R_1$ robot was actuated using only one SMA (left/rear). Consequently, the robot cannot transition into state $(01)$ or between $(00)$ and $(11)$ independently (only via state $(10)$). Thus, the state transition reward matrix for this robot is modified as follows
\begin{equation}
T_{l.o.a.} = 
\begin{array}{c@{}l@{}}
\ & \begin{array}{cccc}
\hspace{8pt} (00) & \hspace{-5pt}(01) & \hspace{-5pt}(10) & \hspace{-5pt}(11)
\end{array}  \\
\begin{array}{c}
(00)\\
(01)\\
(10)\\
(11)
\end{array} %
  & \left[\begin{array}{cccc}
\phantom{-}0  & -1\tikzmark{topA}  & \phantom{-}1  &  \phantom{-}\tikzmark{leftCA}0\tikzmark{rightCA}\\
\tikzmark{leftA}\phantom{-}1 & \phantom{-}0 & \phantom{-}0 & -1\tikzmark{rightA}\\
-1 &\phantom{-}0 & \phantom{-}0 & \phantom{-}1\\
\phantom{-}\tikzmark{leftCB}0\tikzmark{rightCB} & \phantom{-}0\tikzmark{bottomA} & \phantom{-}0 & \phantom{-}0
\end{array}\right]\\
\end{array}
\end{equation}
\noindent 
\DrawLineV[red, ultra thick, opacity=0.5]{topA}{bottomA}
\DrawLineH[red, ultra thick, opacity=0.5]{leftA}{rightA}
\DrawLineH[red, ultra thick, opacity=0.5]{leftCA}{rightCA}
\DrawLineH[red, ultra thick, opacity=0.5]{leftCB}{rightCB}
where the red strikeouts infer the forbidden state transitions. The optimization proceeds in the same way without anything extra being learned. For $l_{max}\leq 5$ the two control sequences for $+X$ translations are $C_{L1}=\{(00) \rightarrow (10) \rightarrow (11) \rightarrow (10) \rightarrow (00)\}$ and $C_{L2} = \{(10) \rightarrow (11) \rightarrow (10)\}$, both resulting in $+1$ resultant  translation.
%
%As observable from the experiments, same control framework is applicable to all three robots where the robot-environment interaction at finite contact points is discretized as two states. The framework is also extendible to loss of limb/actuation scenario. Additionally, the state based control helped in compensation of time dependent inconsistencies in actuating performance of SMA actuators as the independent control parameter is the robot state.

%%%%%%%%%%%%%%%%%%%%%%%%%%%%%%%%%%%%%%%%%%%%%%%%%%%%%%%%%%%%%%%%%%%%
%%%%%%%%%%%%%%%%%%%%%%% CONCLUSION %%%%%%%%%%%%%%%%%%%%%%%%%%%%%%%%%%
\section{Conclusion}
%%%%%%%%%%%%%%%%%%%%%%%%%%%%%%%%%%%%%%%%%%%%%%%%%%%%%%%%%%%%%%%%%%%%
A small, lightweight (3-3.6gms), soft material modular robotic platform has been developed to explore new approaches of soft robot control. The robots are printed on a multi-material 3D printer making their manufacturing and assembly fast and cheap. Inspired by a biological example, the robot is a simple deformable body actuated by muscle-like SMA coils and interacting with the environment at finite discrete points. Instead of building a directly actuated mechanism for controlling grip, friction is manipulated using two materials with different properties. The robot alters its frictional interaction with the environment through changes in shape to produce locomotion.

Control is achieved by indirect modeling of the robot-environment interaction using the model-free control framework. Here, the friction interactions (griper-like behavior) are discretized as binary states $0,1$. These binary states allow definition of four robot states and a state transition matrix. The individual robot state transition rewards are learned and stored in the transition matrix. Optimization allows the sequence of state transitions to be calculated to control robot locomotion. The transition matrix is expected to be helpful for robots maneuvering in unstructured and unpredictable environments as it does not depend on the robot dynamics. This matrix may be learned and repopulated using control feedback to tackle unanticipated environmental changes, thus, making the robot more robust. Another advantage of the state-transition matrix is that it avoids the need to model interactions with the surface. The objective of states-transition matrix is to avoid direct modeling of friction or grip. The state-based discretization allows the control strategy to compensate for inconsistent SMA actuators and simplifies translation of control schemes to non-SMA based soft robot actuators such as motor-tendon based robots. Furthermore, the presented framework is generic, simplifies control parameters, material and actuator independent and extendable to multiple limbs/actuators. 

Experiments demonstrate how this framework can be applied. Three different types of robots generated control sequences that produced anteriograde and retrograde waves equivalent to those seen in soft moving animals \cite{trueman_locomotion_1975}. The approach is also expected to apply to robots moving in changing conditions or when the robot itself undergoes a loss. This was illustrated by turning one of  the actuators off and identifying two control sequences that produce forward locomotion. These experiments also showed that the contact angle $\psi$ (equivalently, state), is sufficient for controlling actuation instead of the actuation time ($T_{(active)}$). 

These bio-inspired robots are simple but powerful platforms to explore new mechanisms of locomotion control in soft devices. They can be easily developed into more structurally complex and capable robots by adding additional beams (analogous to limbs) to produce shapes like the letters Y and X, or increasingly radial configurations such as a starfish. The control framework can also be expanded for planar locomotion in which the robot-environment interaction is a vector of both changes in position and orientation. These studies are underway to explore the advantages and limitations of using a model-free control approach for soft robots in complex environments. 
%%%%%%%%%%%%%%%%%%%%%%%%%%%%%%%%%%%%%%%%%%%%%%%%%%
\section{Acknowledgement}
This work was funded in part by the National Science Foundation grant IOS-1050908 to
Barry Trimmer and National Science Foundation Award DBI-1126382.
%%%%%%%%%%%%%% Bibligography %%%%%%%%%%%%%%%%%%%%%
\bibliographystyle{ieeetr}
\bibliography{softrefs}
\end{document}

%% file: usercommands.tex
\usepackage{graphicx}
\usepackage{amssymb}
\usepackage{multirow} 
\usepackage{collcell}
\usepackage{xstring}

\usepackage{citesort}
\newif\ifconvert
\converttrue
\usepackage{subfig}
\usepackage{array,booktabs}
\newcolumntype{C}[1]{>{\centering\let\newline\\\arraybackslash\hspace{0pt}}m{#1}}

\usepackage{tikz}
\usetikzlibrary{calc}

\newcommand{\tikzmark}[1]{\tikz[overlay,remember picture] \node (#1) {};}
\newcommand{\DrawLineH}[3][]{%
    \begin{tikzpicture}[overlay,remember picture]
        \draw [#1] ($(#2)+(0,0.6ex)$) -- ($(#3)+(0,0.6ex)$);
    \end{tikzpicture}%
}
\newcommand{\DrawLineV}[3][]{%
    \begin{tikzpicture}[overlay,remember picture]
        \draw [#1] ($(#2)+(-0.5ex,1ex)$) -- ($(#3)+(-0.5ex,-0.5ex)$);
    \end{tikzpicture}%
}

%% file: TRA-v2.0.bbl
\begin{thebibliography}{10}

\bibitem{vogel_comparative_2013}
S.~Vogel, {\em Comparative biomechanics: life's physical world}.
\newblock Princeton University Press, 2013.

\bibitem{wang_locomotion_2014}
W.~Wang, J.-Y. Lee, H.~Rodrigue, S.-H. Song, W.-S. Chu, and S.-H. Ahn,
  ``Locomotion of inchworm-inspired robot made of smart soft composite
  ({SSC}),'' {\em Bioinspiration \& Biomimetics}, vol.~9, p.~046006, Dec. 2014.

\bibitem{lin_goqbot:_2011}
H.-T. Lin, G.~G. Leisk, and B.~Trimmer, ``{GoQBot}: a caterpillar-inspired
  soft-bodied rolling robot,'' {\em Bioinspiration \& Biomimetics}, vol.~6,
  p.~026007, June 2011.

\bibitem{seok_peristaltic_2010}
S.~Seok, C.~D. Onal, R.~Wood, D.~Rus, and S.~Kim, ``Peristaltic locomotion with
  antagonistic actuators in soft robotics,'' in {\em {IEEE} {International}
  {Conference} on {Robotics} and {Automation}}, pp.~1228--1233, 2010.

\bibitem{laschi_design_2009}
C.~Laschi, B.~Mazzolai, V.~Mattoli, M.~Cianchetti, and P.~Dario, ``Design of a
  biomimetic robotic octopus arm,'' {\em Bioinspiration \& Biomimetics},
  vol.~4, no.~1, p.~015006, 2009.

\bibitem{hannan_kinematics_2003}
M.~W. Hannan and I.~D. Walker, ``Kinematics and the {Implementation} of an
  {Elephant}'s {Trunk} {Manipulator} and {Other} {Continuum} {Style}
  {Robots},'' {\em Journal of Robotic Systems}, vol.~20, no.~2, pp.~45--63,
  2003.

\bibitem{quinn_parallel_2003}
R.~D. Quinn, G.~M. Nelson, R.~J. Bachmann, D.~A. Kingsley, J.~T. Offi, T.~J.
  Allen, and R.~E. Ritzmann, ``Parallel {Complementary} {Strategies} for
  {Implementing} {Biological} {Principles} into {Mobile} {Robots},'' {\em The
  International Journal of Robotics Research}, vol.~22, pp.~169--186, Mar.
  2003.

\bibitem{pfeifer_challenges_2012}
R.~Pfeifer, M.~Lungarella, and F.~Iida, ``The challenges ahead for bio-inspired
  `soft' robotics,'' {\em Communications of the ACM}, vol.~55, no.~11,
  pp.~76--87, 2012.

\bibitem{hirose_biologically_2004}
S.~Hirose and M.~Mori, ``Biologically inspired snake-like robots,'' in {\em
  {IEEE} {International} {Conference} on {Robotics} and {Biomimetics}},
  pp.~1--7, 2004.

\bibitem{chirikjian_kinematics_1995}
G.~Chirikjian and J.~Burdick, ``The kinematics of hyper-redundant robot
  locomotion,'' {\em IEEE Transactions on Robotics and Automation}, vol.~11,
  pp.~781--793, Dec. 1995.

\bibitem{ostrowski_geometric_1996}
J.~Ostrowski and J.~Burdick, ``The {Geometric} {Mechanics} of {Undulatory}
  {Robotic} {Locomotion},'' {\em The International Journal of Robotics
  Research}, vol.~17, pp.~683--701, 1996.

\bibitem{wright_design_2007}
C.~Wright, A.~Johnson, A.~Peck, Z.~McCord, A.~Naaktgeboren, P.~Gianfortoni,
  M.~Gonzalez-Rivero, R.~Hatton, and H.~Choset, ``Design of a modular snake
  robot,'' in {\em {IEEE}/{RSJ} {International} {Conference} on {Intelligent}
  {Robots} and {Systems}}, pp.~2609--2614, Oct. 2007.

\bibitem{maladen_undulatory_2009}
R.~D. Maladen, Y.~Ding, C.~Li, and D.~I. Goldman, ``Undulatory {Swimming} in
  {Sand}: {Subsurface} {Locomotion} of the {Sandfish} {Lizard},'' {\em
  Science}, vol.~325, pp.~314--318, July 2009.

\bibitem{saranli_rhex:_2001}
U.~Saranli, M.~Buehler, and D.~E. Koditschek, ``{RHex}: {A} {Simple} and
  {Highly} {Mobile} {Hexapod} {Robot},'' {\em The International Journal of
  Robotics Research}, vol.~20, pp.~616--631, July 2001.

\bibitem{shepherd_multigait_2011}
R.~F. Shepherd, F.~Ilievski, W.~Choi, S.~A. Morin, A.~A. Stokes, A.~D. Mazzeo,
  X.~Chen, M.~Wang, and G.~M. Whitesides, ``Multigait soft robot,'' {\em
  Proceedings of the National Academy of Sciences}, vol.~108, pp.~20400--20403,
  Dec. 2011.

\bibitem{umedachi_highly_2013}
T.~Umedachi, V.~Vikas, and B.~A. Trimmer, ``Highly {Deformable} 3-{D} {Printed}
  {Soft} {Robot} {Generating} {Inching} and {Crawling} {Locomotions} with
  {Variable} {Friction} {Legs},'' in {\em {IEEE}/{RSJ} {International}
  {Conference} on {Intelligent} {Robots} and {Systems}}, 2013.

\bibitem{autumn_frictional_2006}
K.~Autumn, A.~Dittmore, D.~Santos, M.~Spenko, and M.~Cutkosky, ``Frictional
  adhesion: a new angle on gecko attachment,'' {\em Journal of Experimental
  Biology}, vol.~209, no.~18, pp.~3569--3579, 2006.

\bibitem{creton_sticky_2007}
C.~Creton and S.~Gorb, ``Sticky feet: from animals to materials,'' {\em MRS
  Bulletin}, vol.~32, no.~06, pp.~466--472, 2007.

\bibitem{gorb_attachment_2001}
S.~Gorb, {\em Attachment devices of insect cuticle}.
\newblock Springer, 2001.

\bibitem{mezoff_biomechanical_2004}
S.~Mezoff, N.~Papastathis, A.~Takesian, and B.~A. Trimmer, ``The biomechanical
  and neural control of hydrostatic limb movements in {Manduca} sexta,'' {\em
  Journal of Experimental Biology}, vol.~207, no.~17, pp.~3043--3053, 2004.

\bibitem{belanger_combined_2000}
J.~H. Belanger and B.~A. Trimmer, ``Combined kinematic and electromyographic
  analyses of proleg function during crawling by the caterpillar {Manduca}
  sexta,'' {\em Journal of Comparative Physiology A}, vol.~186, no.~11,
  pp.~1031--1039, 2000.

\bibitem{radhakrishnan_locomotion:_1998}
V.~Radhakrishnan, ``Locomotion: {Dealing} with friction,'' {\em Proceedings of
  the National Academy of Sciences}, vol.~95, pp.~5448--5455, May 1998.

\bibitem{kim_soft_2013}
S.~Kim, C.~Laschi, and B.~Trimmer, ``Soft robotics: a bioinspired evolution in
  robotics,'' {\em Trends in Biotechnology}, vol.~31, pp.~287--294, May 2013.

\bibitem{ohalloran_review_2008}
A.~OHalloran, F.~OMalley, and P.~McHugh, ``A review on dielectric elastomer
  actuators, technology, applications, and challenges,'' {\em Journal of
  Applied Physics}, vol.~104, no.~7, pp.~071101--071101, 2008.

\bibitem{carpi_biomedical_2009}
F.~Carpi, E.~Smela, and {others}, ``Biomedical {Applications} of
  {Electroactive} {Polymer} {Actuators},'' 2009.

\bibitem{cianchetti_new_2009}
M.~Cianchetti, V.~Mattoli, B.~Mazzolai, C.~Laschi, and P.~Dario, ``A new design
  methodology of electrostrictive actuators for bio-inspired robotics,'' {\em
  Sensors and Actuators B: Chemical}, vol.~142, no.~1, pp.~288--297, 2009.

\bibitem{daerden_pneumatic_2002}
F.~Daerden and D.~Lefeber, ``Pneumatic artificial muscles: actuators for
  robotics and automation,'' {\em European Journal of Mechanical and
  Environmental Engineering}, vol.~47, no.~1, pp.~11--21, 2002.

\bibitem{wehner_experimental_2012}
M.~Wehner, Y.-L. Park, C.~Walsh, R.~Nagpal, R.~J. Wood, T.~Moore, and
  E.~Goldfield, ``Experimental characterization of components for active soft
  orthotics,'' in {\em {IEEE} {RAS} \& {EMBS} {International} {Conference} on
  {Biomedical} {Robotics} and {Biomechatronics}}, pp.~1586--1592, 2012.

\bibitem{van_griethuijsen_locomotion_2014}
L.~van Griethuijsen and B.~Trimmer, ``Locomotion in caterpillars,'' {\em
  Biological Reviews}, 2014.

\bibitem{_http://www.stratasys.com/_????}
{\em http://www.stratasys.com/}.

\bibitem{_http://www.toki.co.jp/biometal/english/contents.php_????}
{\em http://www.toki.co.jp/biometal/english/contents.php}.

\bibitem{wayman_shape_1993}
C.~Wayman, ``Shape {Memory} {Alloys},'' {\em MRS Bulletin}, vol.~18,
  pp.~49--56, Apr. 1993.

\bibitem{asbeck_scaling_2006}
A.~T. Asbeck, S.~Kim, M.~R. Cutkosky, W.~R. Provancher, and M.~Lanzetta,
  ``Scaling {Hard} {Vertical} {Surfaces} with {Compliant} {Microspine}
  {Arrays},'' {\em The International Journal of Robotics Research}, vol.~25,
  no.~12, pp.~1165--1179, 2006.

\bibitem{webster_design_2010}
R.~J. Webster and B.~A. Jones, ``Design and {Kinematic} {Modeling} of
  {Constant} {Curvature} {Continuum} {Robots}: {A} {Review},'' {\em The
  International Journal of Robotics Research}, vol.~29, pp.~1661--1683, Nov.
  2010.

\bibitem{walker_continuum_2011}
I.~Walker, ``Continuum robot appendages for traversal of uneven terrain in in
  situ exploration,'' in {\em {IEEE} {Aerospace} {Conference}}, pp.~1--8, Mar.
  2011.

\bibitem{cianchetti_design_2011}
M.~Cianchetti, A.~Arienti, M.~Follador, B.~Mazzolai, P.~Dario, and C.~Laschi,
  ``Design concept and validation of a robotic arm inspired by the octopus,''
  {\em Materials Science and Engineering: C}, vol.~31, pp.~1230--1239, Aug.
  2011.

\bibitem{duriez_control_2013}
C.~Duriez, ``Control of elastic soft robots based on real-time finite element
  method,'' in {\em {IEEE} {International} {Conference} on {Robotics} and
  {Automation}}, pp.~3982--3987, May 2013.

\bibitem{trueman_locomotion_1975}
E.~R. Trueman, {\em The {Locomotion} of {Soft}-{Bodied} {Animals}.}
\newblock American Elsevier Publishing Co., 1975.

\end{thebibliography}
